\documentclass[10pt,twocolumn,letterpaper]{article}
\usepackage{cvpr}
\usepackage{times}
\usepackage{subfigure}
\usepackage{epsfig}
\usepackage{graphicx}
\usepackage{amsmath}
\usepackage{amssymb}
\usepackage{amsthm}
\usepackage[pagebackref=true,breaklinks=true,letterpaper=true,colorlinks,bookmarks=false]{hyperref}
\usepackage[linesnumbered,boxed]{algorithm2e}
\renewcommand{\vec}[1]{\mbox{\boldmath ${#1}$}}
\newcommand{\Matrix}[1]{\mbox{\boldmath ${#1}$}}

\cvprfinalcopy
\begin{document}
\def\cvprPaperID{481} 
\def\httilde{\mbox{\tt\raisebox{-.5ex}{\symbol{126}}}}
\title{Cross-view Action Modeling, Learning and Recognition}
\author{Jiang Wang\textsuperscript{1} \quad Xiaohan Nie\textsuperscript{2} \quad Yin Xia\textsuperscript{1}  \quad  Ying Wu\textsuperscript{1} \quad Song-Chun Zhu\textsuperscript{2} \\
\textsuperscript{1}Northwestern University \quad \textsuperscript{2}University of California at Los Angles
}
\maketitle
 \thispagestyle{empty}

\begin{abstract}
Existing methods on video-based action recognition are generally view-dependent, i.e., performing recognition from the same views seen in the training data. We
present a novel multiview spatio-temporal AND-OR graph (MST-AOG) representation for  cross-view action recognition, i.e., the recognition is
performed on the video from an unknown and unseen view. As a compositional model, MST-AOG compactly represents the hierarchical combinatorial
structures of cross-view actions by explicitly modeling the geometry, appearance and motion variations. This paper proposes effective methods to
learn the structure and parameters of MST-AOG. The inference based on MST-AOG enables action recognition from novel views. The training of
MST-AOG takes advantage of the 3D human skeleton data obtained from
Kinect cameras to avoid annotating enormous multi-view video frames, which is
error-prone and time-consuming, but the recognition does not need 3D information and is based on 2D video input. A new Multiview Action3D dataset has
been created and will be released. Extensive experiments have demonstrated that this new action representation significantly improves the
accuracy and robustness for cross-view action recognition on 2D videos.
\end{abstract}

\section{Introduction}
In the literature of video-based action recognition, most existing methods recognize actions from the view that is more or less the same as
the training videos~\cite{Ikizler-Cinbis2012}. Their general limitation is the unpredictable  performance in the
situation where the actions need to be recognized from a novel view. As the visual appearances are very different from
different views, and it is very difficult to find view-invariant features.
 Therefore, it is desirable  to build models for cross-view action recognition, i.e., recognizing video actions from views that are unseen in the training videos. Despite some recent attempts~\cite{Liu2011, Junejo2008}, this problem
has not been well explored.

One possible approach is to  {\em enumerate} a sufficiently large number of views
and build dedicated feature and classifier for each view. This
approach is too time consuming, because it requires annotating a large number of videos for all views
multiplied by all action categories. Another possible approach is to
{\em interpolate} across views via transfer learning~\cite{Liu2011}. This method
learns a classifier from one view, and adapts
the classifiers to new views. The performance of this approach is largely
limited by the discrimination power of the local spatio-temporal
features in practice.

In this paper, we approach this problem from a new perspective: creating a generative cross-view video action representation by exploiting the
compositional structure in spatio-temporal patterns and geometrical relations among views. We call this model multiview spatio-temporal
AND-OR graph model (MST-AOG), inspired by the expressive power of AND-OR graphs in object modeling~\cite{Sia}. This model
includes multiple layers of nodes, creating a hierarchy of composition at various semantic levels, including {\em actions}, {\em poses}, {\em
views}, {\em body parts} and {\em features}. Each node represents a conjunctive or disjunctive composition of its children nodes.
The leaf nodes are appearance and motion features that ground the model. An important feature of the MST-AOG model is that the grounding does not have to be at
the lowest layer (as in conventional generative models), but can be made at upper layers to capture low resolution spatial and temporal features.
This compositional representation models geometry, appearance, and motion properties for actions. Once the model is learned, the
inference process facilitates cross-view pose detection and action classification.

The AND/OR structure of this MST-AOG model is simple, but the major challenges lie in the learning of geometrical relations among different
views. This paper proposes novel solutions to address this
difficult issue. To learn  the multiple-view structure, we take advantage of the 3D human skeleton produced by Kinect sensors as
the 3D pose annotation. This 3D skeleton information is only available in training, but not used for cross-view action recognition. The
projection of the 3D poses enables explicit modeling of the 2D views. Our model uses a set of discrete views in training to interpolate arbitrary
novel views in testing. The appearances and motion are learned from the multiview training video and the 3D pose skeletons.

To learn the  multiple-pose structure, we design a new discriminative data mining method to automatically discover the frequent and
discriminative poses. This data-driven method provides a very effective way to learn the structure for the action nodes.
Since this hierarchical structure enables information sharing (e.g., different {\em view} nodes share certain body {\em part} nodes), MST-AOG
largely reduces the enormous demands on data annotation, while improving the accuracy and robustness of cross-view action recognition, as
demonstrated in our extensive experiments.

%

\section{Related Work and Our Contributions}
The literature on action recognition can be roughly divided into the following categories:

{\bf Local feature-based methods.} Action recognition methods can be based on the bag-of-words representation of local features, such as
HOG~\cite{Dalal} or HOF~\cite{Laptev2008} around spatio-temporal interest points~\cite{Laptev2005}.  Transfer learning-bsaed cross-view action recognition
methods~\cite{farhadi2008learning, Li2012, zhang2013cross} are based on local appearance features. Hankelet~\cite{li2012cross} represents actions
with the dynamics of short tracklets, and achieves cross-view action recognition by finding the Hankelets that are relative invariant to viewpoint changes.
Self temporal similarity~\cite{Junejo2008} characterizes actions with
temporal self-similarities for cross-view action recognition. These methods work well on simple action classification,
but they usually lack discriminative power to deal with more complex actions.

{\bf 2D Pose-based methods.} Recently, human pose estimation from a single image has make great progresses~\cite{Yang2011}. There is emerging
interest in exploiting human pose for action recognition. Yao et al.~\cite{Yao2012} estimates the 2D poses from the images, and matches the estimated
poses with a set of representative poses. Yao et al.~\cite{YaoB,yao2011deformable}  developed spatio-temporal AND-OR graph to model the
spatio-temporal structure of the poses in an actions. Desai et al.~\cite{Desai2012} learns a deformable part model (DPM)~\cite{Felzenszwalb2010a} that estimates both
human poses and object locations. Maji et al.~\cite{Maji2011a} uses the activations of {\em poselets}, which is is a set of pose detectors.
Ikizler-Cinbis et al.~\cite{Ikizler-Cinbis2012} learns the pose
classifier from web images. \cite{yao2012coupled} proposes a coupled action recognition and
pose estimation method by formulating pose estimation as an
optimization over a set of action-specific manifold.  In general, these methods were not specifically designed to handle cross-view actions. In contrast, this paper presents a new multi-view video action recognition approach.

{\bf 3D skeleton-based methods.} Pose-based action recognition generally needs a large amount of annotated poses from images. Recently, the
development of depth cameras offers a cost-effective method to track 3D human poses~\cite{Shotton2011}. Although the tracked 3D skeletons are
noisy, it has been shown that they are useful to achieve good results in recognizing fine-grained actions~\cite{Wang2012}. In addition, the 2D
DPM model can be extended to 3D~\cite{Fidler,Liebelt2010} to facilitate multi-view object detection. Parameswaran~\cite{parameswaran2006view}
proposes view-invariant canonical body poses and trajectories in 2D invariant space. In this paper, our proposed
method uses the tracked 3D skeleton as supervision in training, but it stands out from other skeleton-based method because it does not need 3D
skeletons inputs for action recognition in testing.

In comparison with the literature, this paper makes the following contributions:
\vspace{-6pt}
\begin{itemize}\addtolength{\itemsep}{-0.5\baselineskip}
\item The proposed MST-AOG model is a compact but expressive
multi-view action representation that unifies the modeling of
geometry, appearance and motion.
\item  Once trained, this MST-AOG model only needs 2D video input to recognize actions from novel views.
\item To train this MST-AOG model, we provide new and effective methods to learn its parameters, as well as mining its structure to enable
effective part sharing.
\end{itemize}\vspace{-6pt}

\section{Multi-view Spatio-Temporal AOG}
\subsection{Overview}
 Being a multi-layer hierarchical compositional model,
the proposed multiview spatio-temporal AND-OR graph (MST-AOG) action representation is able to compactly accommodate the combinatorial
configurations for cross-view action modeling. It consists of AND, OR and
leaf nodes at various layers, and each node is associated with a score computed from its children. An AND node models the conjunctive
relationship of its children nodes, and its score takes the summation over those of its children. An OR node captures the disjunctive
relationship or the mixture of possibilities of its children node, and its score takes the maximum over its children. A leaf node is observable
and is associated with evidence, and thus grounds the model.

\begin{figure}
  \begin{center}
     \includegraphics[width = 8cm]{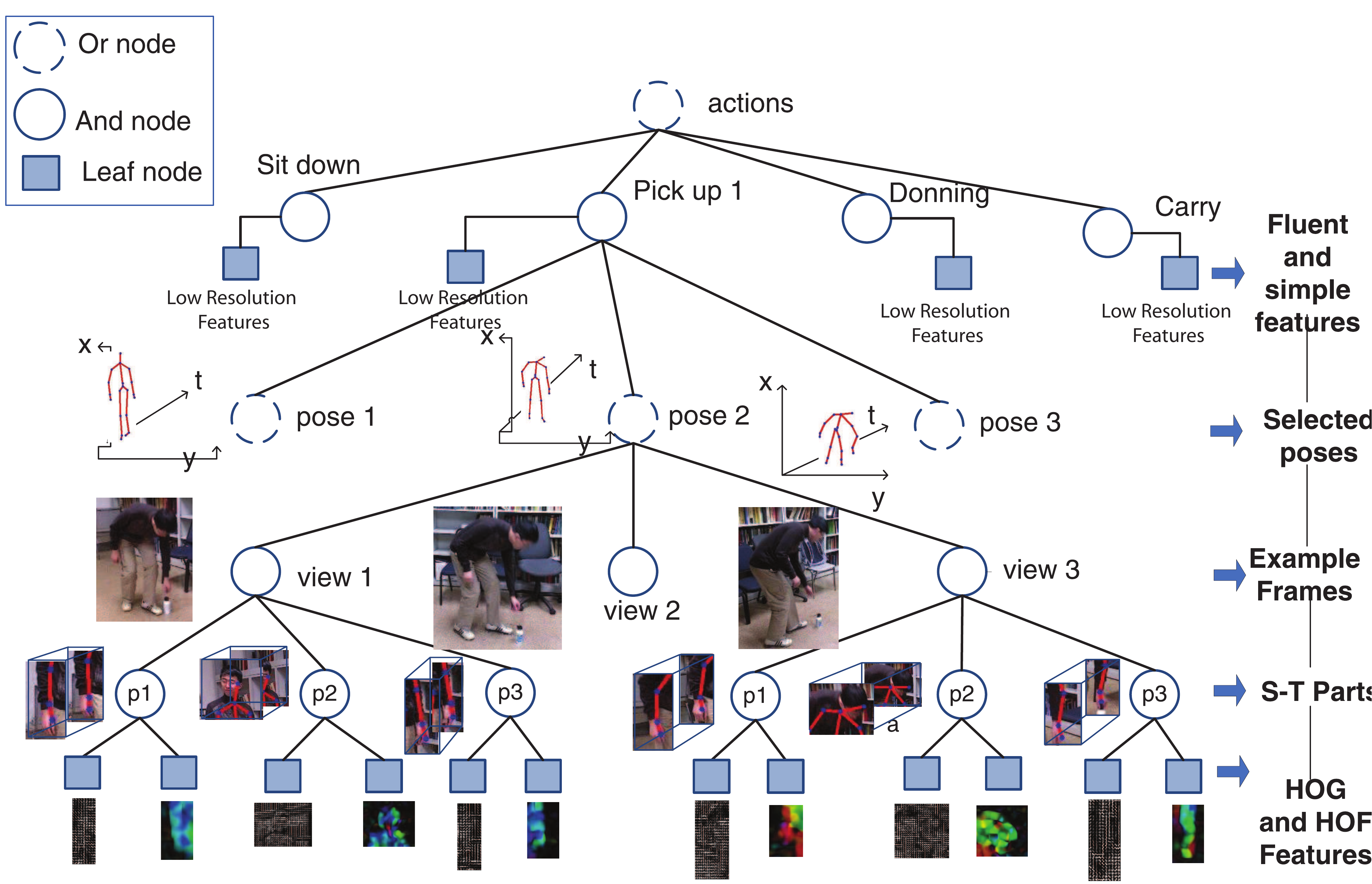}
  \caption{The MST-AOG action representation. The geometrical relationship of
  the parts in different views are modeled jointly by projecting the 3D poses into
  the given view, see Fig.~\ref{fig:pose}. The parts are discriminately mined and shared for all the actions.} \vspace{-24pt}
  \label{fig:and_or}
  \end{center}
\end{figure}

The structure of the proposed MST-AOG model is shown in Fig.~\ref{fig:and_or}. The {\em root} node is an OR node representing the mixture of the
set of all actions. We regard an action as a sequence of discriminative 3D poses. A 3D pose exhibits a mixture of its projections on a set of 2D
views. A 2D view includes a set of spatio-temporal parts, and each part is associated with its appearance and motion features. Thus, the {\em
action} nodes, {\em view} nodes and {\em part} nodes are AND nodes, and {\em pose} nodes are OR nodes. We will discuss the scores and parameters
for these nodes in the following subsections.

The strong expressive power of an AOG~\cite{Sia} lies in the structure
of layered conjunctive and disjunctive compositions.
Moreover, MST-AOG shares the part nodes across different views via interpolation. An example will be given when discussing the
{\em action} node in Sec.~\ref{sec:action-node}.

\subsection{Pose/View Nodes and 3D Geometry}\label{sec:geometry}

\begin{figure}
  \begin{center}
    \includegraphics[width=5.2cm]{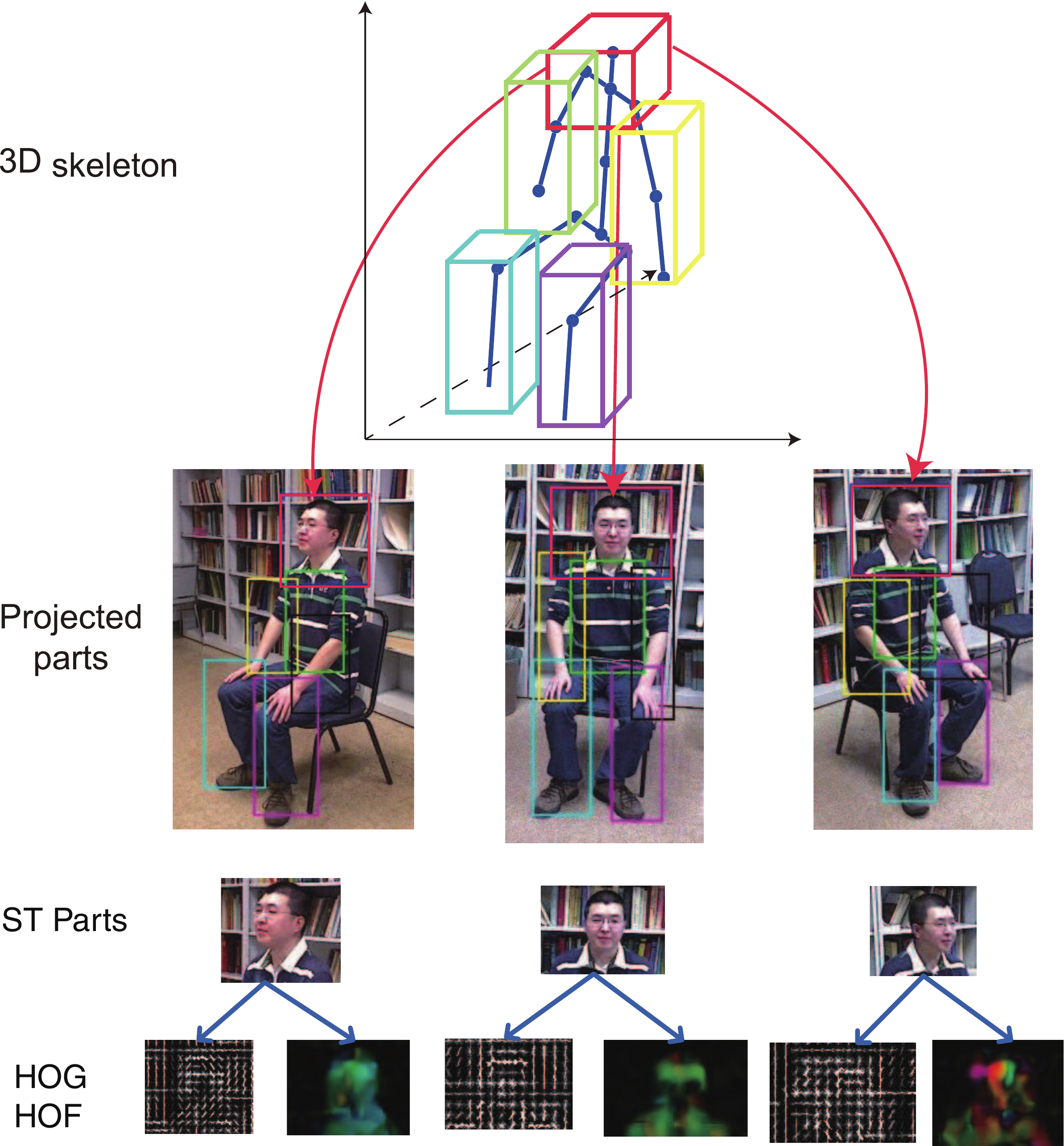}
  \end{center}
  \caption{3D parts and projected parts in different views.}\vspace{-18pt}
  \label{fig:pose}
\end{figure}

To handle multi-view modeling, we introduce {\em pose} and {\em view} nodes. A {\em pose} node is an OR node that models the association of
spatio-temporal patterns to a 3D pose projected to various views (each of which is a {\em view} node). For each {\em view} node, it captures the
AND relationship of a number of parts (i.e., the limb of the human). Each {\em part} node captures its visual appearance and motion features
under a specific view $\theta$. Specifically, we use a star-shaped model for the dependencies among body parts, inspired by
DPM~\cite{Felzenszwalb2010a}, as  Fig.~\ref{fig:pose} shows.
Their 2D locations are denoted by $\mathcal V = \{ \vec v_0, \vec v_1,
\cdots, \vec v_N\}$, where $\vec v_0$ is for the root part (the whole pose). Denote by $I$
the image frame. We define the score associated with the $i$-th {\em part} node to be $S_{\cal R} (\vec v_i, I, \theta)$ (details will be
provided in Sec.~\ref{sec:appearance_motion}).

Two factors contribute to the score of a {\em view} node: the score of its children {\em part} nodes $S_{\cal R} (\vec v_i, I, \theta)$ and the
spatial regularization among them $S_i(\vec v_0, \vec v_i, \theta)$ that specifies the spatial relationship between the root part and each child
part. Such spatial regularization measures the compatibility among the parts from view $\theta$ (we only consider the rotation angle, details will follow). In view of this, the total compatibility score of a {\em view} node is written as:
\begin{equation}\label{eq:score_dpm}
\begin{small}
  S_{\cal V}
(\vec v_0, \theta) = \sum_{i=0}^{N} S_{\cal R} (\vec v_i,I,\theta) + \sum_{i=1}^{N} S_i(\vec v_0, \vec v_i, \theta)
\end{small}
\end{equation}
where $\vec v_i$ is the location of the part $i$, and $\theta$ is the view.

The 2D global location of a 2D pose is set to be the location of the root part, i.e.,$\vec v_0$. As the {\em pose} node is an OR node, the score
for a {\em pose} node is computed by maximizing the scores from its children {\em view} nodes:
\begin{equation}\label{eq:view_node}
 S_{\cal P}(\vec v_0) = \max_{\theta} S_{\cal V}(\vec v_0, \theta)
\end{equation}

The evaluation of the spatial regularization of the parts needs a special treatment, because a {\em pose} node represents a 3D pose and it can
be projected to different views to lead to different part relationships explicitly, as illustrated in Fig. \ref{fig:pose}.

The 3D geometrical relationship of the parts can be modeled as the 3D offsets of the $i$-th part with respect to the root part. Each offset can
be modeled as a 3D Gaussian distribution with the mean $\vec \mu_i$ as well as diagonal covariance matrix $\Matrix \Sigma_i$.
\begin{equation}\label{eq:displacement}
 \log P(\Delta \vec p_i) \propto -\frac 1 2 \left( \Delta \vec p_i - \vec \mu_i\right)^T
   \Matrix \Sigma_i^{-1} \left( \Delta \vec p_i - \vec \mu_i\right)
\end{equation}
where $\Delta \vec p_i = (\Delta x_i, \Delta y_i, \Delta z_i)$ is the 3D offset between the part $i$ and the root part. Here $\vec \mu_i$ can be
estimated using the 3D skeleton data, and $\Matrix \Sigma_i$ will be learned (in Sec.~\ref{sec:learning}).

The distribution of the 3D part offsets is projected to 2D for a given view. Here we assume scaled orthographic projections: $\Matrix Q_i^\theta$
\begin{equation}\label{eq:orthographic_projection}
\Matrix Q_i^\theta =
\begin{bmatrix}
k_1\cos\theta  & 0 & -k_1 \sin\theta\\
0 & k_2 &0 &  \\
\end{bmatrix}
\end{equation}
where $\theta$ is a rotation angle of the view, and $k_1$ and $k_2$ are the scale factors for two image axes. In training, we take advantage of
the 3D skeleton data from Kinect cameras. Since we have the ground truth 3D (from 3D skeleton data) and 2D (from multiview videos) locations in
our training data, these parameters can be easily estimated. The
orthographic projection approximation works well in practice because the actors are sufficiently far away
from the camera when performing actions. Since $\Matrix Q_i^\theta$ is a linear transform, the resulting projected 2D offset distribution is also
a Gaussian distribution, with mean $\vec \mu_i^\theta = \Matrix Q_i^\theta \vec v_i$ and covariance matrix $\Matrix \Sigma_i^\theta = \Matrix
Q_i^\theta \Matrix \Sigma (\Matrix Q_i^\theta)^T$. Thus the 2D spatial pairwise relationship score $S_{i}(\vec v_0 , \vec v_i, \theta)$ can be
written as follows:
\begin{equation}\label{eq:score_relationship}
  \begin{small}
  \begin{aligned}
S_{i}(\vec v_0 ,  \vec v_i, \theta) &= ((\Matrix \Sigma_i^\theta)^{-1}_{11}, (\Matrix \Sigma_i^\theta)^{-1}_{22}, (\Matrix \Sigma_i^\theta)^{-1}_{12})^T .\\
& (-\Delta u_i^2, -\Delta v_i^2, -2 \Delta u_i \Delta v_i)
  \end{aligned}
  \end{small}
\end{equation}
where $(\Delta u_i, \Delta v_i) = \vec v_i - \vec v_0 - \vec \mu_i^\theta$ is the 2D deformation between the $i$-th part  and the root part.

This 3D geometrical relationship is shared and learned across different
views. The 2D geometrical relationship of the novel views can be
obtained by projecting the 3D geometrical relationship to the novel views.

\subsection{Part Node and Motion/Appearance} \label{sec:appearance_motion}

The spatio-temporal patterns of a part under a view are modeled as its motion and appearance features. Each {\em part} has an {\em appearance}
node with score $A_i(\vec v_i, I, \theta)$, and a {\em motion} node with score $M_i(\vec v_i, I, \theta)$. They capture the likelihood (or
compatibility) of the appearance and motion of part $i$ located at $\vec v_i$ under view $\theta$, respectively. The score associated with a {\em
part} node is thus written as:
\begin{equation}\label{eq:part_score}
\begin{small}
  S_{\cal R} (\vec v_i, I, \theta) = A_i(\vec v_i, I, \theta) + M_i(\vec v_i, I, \theta).
\end{small}
\end{equation}

We exploit commonly used HOG \cite{Dalal} and HOF \cite{Laptev2008} features to represent the appearance and motion of a given part,
respectively. In order to model the difference and correlation of the appearance and motion for one part in different view, we discretize the
view angle $\theta$ into $M$ discrete bins (each bin corresponds to a
view node), and use exponential interpolation to obtain the appearance and motion features in the view bins.
The appearance score function  $A_i(\vec v_i, I, \theta)$ and motion score function   is defined as
\begin{equation}\label{eq:appearance}
  \begin{small}
  A_i(\vec v_i, I, \theta)= \frac{\sum_{m=1}^{M}{e^{-d^2(\theta, \theta_m)} \vec \phi_{i,m}^T
    \phi(I, \vec v_i, \theta)}}{\sum_{m=1}^{M}{e^{-d^2(\theta, \theta_m)}}}
  \end{small}
\end{equation}
where $e^{-d^2(\theta, \theta_m)}$ is the exponential of angular distance between the view $\theta$ and the view of
bin $m$, $\vec \phi(I, \vec v_i, \theta)$ is the HOG  features at the location $\vec v_i$ in image $I$ under the view
$\theta$. $\vec \phi_{i,m}$  is the HOG  templates of view bin $m$, and need to be learned from the
training data (see Sec.~\ref{sec:learning}). The motion score function $M_i(\vec v_i, I, \theta)$ is defined and learned from HOF features in a similar
way.

Thus, the part node of different nodes are shared across different
views via interpolation. We can learn the appearance/motion of the part nodes for the novel views via interpolation.
\subsection{Action Node}
\label{sec:action-node}

Basically an action consists of a number of $N_{\cal P}$ 3D discriminative poses, but it is insufficient for an {\em action} node to include only
a set of {\em pose} nodes for two reasons. First, when the image resolution of the human subject is low, further decomposing the human into body
parts is not plausible, as detecting and localizing such tiny body parts will not be reliable. Instead, low resolution visual features may allow
the direct detection of rough poses. Suppose we have $N_{\cal L}$ low resolution features,
denoted by $\vec \varphi_i, i = 1, 2, \cdots, N_{\cal L}$. We simply use a linear prediction function $\sum_i^{N_{\cal L}} \vec w_i^T \vec \varphi_i$ to evaluate
low-resolution-feature action prediction score. The weights $\vec w_i$ can be learnt for each low-resolution features.
We use two low-resolution features: intensity histogram and size of the bounding boxes of the foreground.

Therefore, an {\em action} node consists of two kinds of children nodes: a ${N_{\cal P}}$ number of {\em pose} nodes and  a ${N_{\cal L}}$ number
of leaf nodes for low-resolution grounding. The score of an {\em action} node evaluates:
\begin{equation}\label{eq:action_node}
  \begin{small}
  S_{\cal A}(l) = \sum_i^{N_{\cal P}} S_{\cal P}^i(\vec v_0)+ \sum_i^{N_{\cal L}} \vec w_i^T \vec \varphi_i
  \end{small}
\end{equation}
where $S_{\cal P}^i(\vec v_0)$ is the score of the $i$-th {\em pose}
node, and $\vec w_i$ is weights of the low-resolution features to be learned (as discussed
in Sec.~\ref{sec:learning}).

\section{Inference}\label{sec:inference}

Given an input video from a novel view, the inference of MST-AOG calculates the scores of all the nodes so as to achieve cross-view action
classification. Since this MST-AOG model is tree-structured, inference can be done via dynamic programming. The general dynamic programming
process contains bottom-up phase and top-down phase, which is similar to sum-product and max-product algorithm in graphical model.

\subsection{Cross-view Pose Detection}

The states of the {\em pose} nodes, {\em view} nodes, and {\em part} nodes are their locations and scales. The score for a {\em view} node is
defined in Eq. \eqref{eq:score_dpm}, and the score for a {\em pose} node is defined in Eq. \eqref{eq:view_node}. The inference of a {\em pose}
node is simply comparing the scores of all the child {\em view} nodes at each location and scale, and finding  the maximum score.

For a {\em view} node, since the score function  \eqref{eq:score_dpm} is convex, we can maximize the score in terms of the locations of the parts
$\vec v_0, \vec v_1, \cdots, \vec v_N$ very efficiently using distance transform \cite{Felzenszwalb2010a}. The inference step can be achieved by
convolving the input frame and its optical flow with the appearance and motion templates of all the parts from different views and obtain the
response maps. Then for each view bin, we can compute its projected part offset relationship. Using the distance transform, we can efficiently
calculate the response map for the poses under this view bin. This also enables the estimation of the novel view by finding the view bin that has
the largest view score.

\subsection{Action Classification}\label{sec:action_class}

We apply the spatio-temporal pyramid to represent the spatio-temporal relationship of poses and low-resolution features for action recognition. The scores of the
{\em pose} nodes and the {\em low-resolution feature} nodes at different locations and frames constitute a sequence of response maps. We apply the max-pooling
over a spatio-temporal pyramid. The response of a cell in the pyramid is the maximum among all responses in this cell.

We divide one whole video into 3-level pyramid in the spatio-temporal
dimensions. This yields $1 + 8 + 64 = 73$ dimensional vector for each
response map.
 Then, we can use the linear prediction function defined in Eq. \eqref{eq:action_node} to compute the score of an
action. The {\em action} node with the maximum score corresponds to
the predicted action. Although this representation only acts as a rough description of the spatial-temporal relationships between the poses, we find it achieves very
good results on our experiments.

\section{Learning}\label{sec:learning}

The learning process has two tasks. The first is to learn the MST-AOG parameters, e.g., the appearance and motion patterns of each part in the
{\em part} nodes, 3D geometrical relationship in the {\em view} and {\em pose} nodes, and the classification weights in the {\em action} nodes.
The second task is to discover a dictionary of discriminative 3D poses to determine the structure  of the MST-AOG model.

\subsection{Learning MST-AOG Parameters}

Learning MST-AOG parameters for the {\em part} and {\em view} nodes can be formulated as a latent structural SVM problem. The parameters of the
latent SVM include: the variance $\Matrix \Sigma_i$ in Eq.~\eqref{eq:displacement}, the appearance and motion templates $\vec \beta_{i,m}$ and
$\vec \gamma_{i,m}$ in Eq. \eqref{eq:appearance}.

Although we have the non-root part locations and the view available in the training data, since we are more interested in predicting the pose
rather than the precise location of each part and the view, we treat the locations of the parts $\vec v_j$ and the view $\theta$ as latent
variables. And we apply a latent SVM to learn the our model using the labeled location of the parts and the view angle as initialization. This
treatment is more robust to the noise in the training data.

For each example $\vec x_n$, we have its class label $y_n \in \{-1, +1\}$, $n \in \{ 1,2, \cdots, N\}$. The objective function is:
\begin{equation}\label{eq:learning}
  \begin{small}
  \begin{aligned}
    \min_{\vec \beta, \vec \gamma, \Matrix
\Sigma_i} & \frac{1}{2} \|[\beta, \vec \gamma, \Matrix
\Sigma_i]\|_2^2+ C \sum_{n=1}^{N}{\max\left(0, 1- y_n S_{\cal
  P}(\vec v_0: \vec x_n)\right)}
  \end{aligned}
  \end{small}
\end{equation}
where $S_{\cal P}(\vec v_0: \vec x_i)$ is defined in Eq. \eqref{eq:view_node}, which is the total score for example $\vec x_i$.

The learning is done by iterating between optimizing $\beta, \gamma, \Matrix \Sigma_i$, and calculating the part locations and the views of the
positive training data.

For each pose, we use the samples whose distances are less than $\eta$ to this pose in the positive videos as positive examples, and randomly
sample 5000 negative training examples from negative videos. We apply two bootstrapping mining of hard negatives during the learning process. As the
action score Eq. \eqref{eq:action_node} is a linear function, the parameter $\vec w_i$ can be easily learned via a linear
SVM solver.

\subsection{Mining 3D Pose Dictionary}\label{sec:pose_mining}

To learn the structure of the MST-AOG, we propose an effective data mining method to discover the discriminative 3D poses, which are specific
spatial configurations of a subset body parts.

\subsubsection{Part Representation}\label{sec:representation}
The 3D joint positions are employed to characterize the 3D pose of the human body. For a human subject, 21 joint positions are tracked by the
skeleton tracker~\cite{Shotton2011} and each joint $i$ has 3 location coordinates $\vec p_j(t) = (x_j(t), y_j(t), z_j(t))$, a motion vector $\vec
m_j(t) = (\Delta x_j(t), \Delta y_j(t) , \Delta z_j(t))$ as well the visibility label $h_j(t)$ at a frame $t$. $h_j(t) = 1$ indicates that the
$j$-th joint is visible in frame $t$ and $h_j(t) = 0$ otherwise. The location coordinates are normalized so that they are invariant to the
absolute body position, the initial body orientation and the body size. We manually group the joints into multiple parts.

\subsubsection{Part Clustering}
Since the poses in one action are highly redundant, we cluster the examples of each part to reduce the size of the search space, and to enable
part sharing. Let part $k$ be one of the $K$ parts of the person and $\mathcal J_k$ be the set of the joints of this part. For each joint $j \in
\mathcal J_k$ in this part, we have $\vec p_j = (x_j, y_j, z_j)$, $\vec m(j) = (\Delta x, \Delta y, \Delta z)$, and $\vec h_i \in \{ 0, 1 \}$ as
its 3D position, 3D motion and visibility map, respectively. For a certain part, given the 3D joint positions of the two examples $s$ and $r$, we
can define their distance:
\begin{equation}\label{eq:disance}
  \begin{aligned}
D_k(s, r) = &\sum_{j \in \mathcal J_k} (\|\vec p_j^s(t) - \hat{\Matrix S}
\vec p_j^r(t)\|_2^2 \\
& + \|\vec m_j^s(t) - \hat{\Matrix S} \vec
m_j^r(t)\|_2^2 )\left(1 + h_{s,r}(t)\right)
  \end{aligned}
\end{equation}
where $\hat{\Matrix S}$ is a similarity transformation matrix that minimizes the distance between the part $k$ of the example $s$ and the example
$t$. The term $h_{s,r}$ is a penalty term based on the visibility of the joint $j$ in the two examples: $h_{s,r}(j) = a$ if $v_s(j) = v_r(j)$ and
is 0 otherwise. Since this distance is non-symmetric, we use a symmetric distance as the  distance metric: $ \bar D(s,r ) = (D(s,r) + D(r,s))/2$.

Spectral clustering is performed on the distance matrix. We remove the clusters that have too few examples, and use the rest of the clusters as
the candidate part configurations for mining. We denote the set of all candidates part configurations for the part $k$ as: $\mathcal T_k = \{
t_{1k}, t_{2k}, \cdots, t_{N_kk}\}$, where each $t_{ik}$ is called a {\em part item} represented by the average joint positions and motions in
the cluster.

\subsubsection{Mining Representative and Discriminative Poses}
The discriminative power of a single part is usually limited. We need to discover poses (the combinations of the parts) that are discriminative for
action recognition.

For a pose $\cal P$ that contains a set of part items $\mathcal T(\cal P)$, with each part item in this set belonging to different part. we
define the spatial configuration of a poses as the 3D joint positions and motions of all the part items in this pose.

The activation of a pose $\cal P$ with configuration $\vec p_{\cal P}$
in a video $v_i$ can be defined as: $a_{p}(i) = \min_{t} e^{-D(\vec
  p_{\cal P}, \vec p_{\cal P}^t)}$, where $\vec p_{\cal P}^t$ is the 3D joint positions of the poses $\cal P$ in the $t$-th frame of video, and $D(., .)$ is a
distance function defined in Eq.~\eqref{eq:disance}. If very similar poses exist in this video, the activation is high. Otherwise, the
activation is low. One discriminative pose should have large
activation in the videos in a given category, while having low activation vector in
other categories. We define the support of the pose $\cal P$ for category $c$ as:$Supp_p(c) = \frac{\sum_{c_i = c} a_p(i)}{\sum_{c_i = c} 1}$,
where $c_i$ the category label of video $v_i$, and the discrimination of the poses $p$ as: $\text{Disc}_p(c) = \frac{Supp_p(c)}{\sum_{c' \neq
c} Supp_p(c')}$.

We would like to discover the poses with large support and discrimination. Since adding one part item into a pose
always creates another pose with lower support, i.e., $Supp_{\cal P}(c) <
Supp_{\cal P'}(c)$ if $\mathcal T(\cal P) \supset \mathcal T(\cal P')$.
Thus we can use the Aprior-like algorithm  to find the discriminative
poses.  In this algorithm, we remove the non-maximal poses
from the discriminative pose pool. For a pose $\cal P$, if there
  exist a pose $\cal P'$ such that $\mathcal T(\cal P) \subset \mathcal  T(\cal P')$ and both $\cal P$ and $\cal P'$ are in the set of discriminative and
  representative poses, then $\cal P$ is a non-maximal pose.

This algorithm usually produces an excessive large number of poses, we
prune the sets of discriminative poses with the following criteria.
Firstly, we  remove poses that are similar to each other. This can be
  modeled as a set-covering problem, and can be solved with a greedy
  algorithm. We choose a pose $\cal P$ with highest discrimination, and
  remove the poses whose distance is less than a given
  threshold. Secondly, we  remove the poses with small validation scores for the
detectors trained for  these poses.

\section{Experiments}
We evaluate the proposed method on two datasets: the Multiview Action3D
Dataset, collected by ourselves and  the MSR-DailyActivity3D dataset~\cite{Wang2012}.

In all our experiments, we only use the videos from a single unknown view for testing, and do not use the skeleton information or the videos from
multiple views.

\subsection{Northwestern-UCLA Multiview Action3D Dataset}\label{sec:action3D}
\begin{figure}
  \begin{center}
\subfigure[]{
  \includegraphics[width=3.0cm]{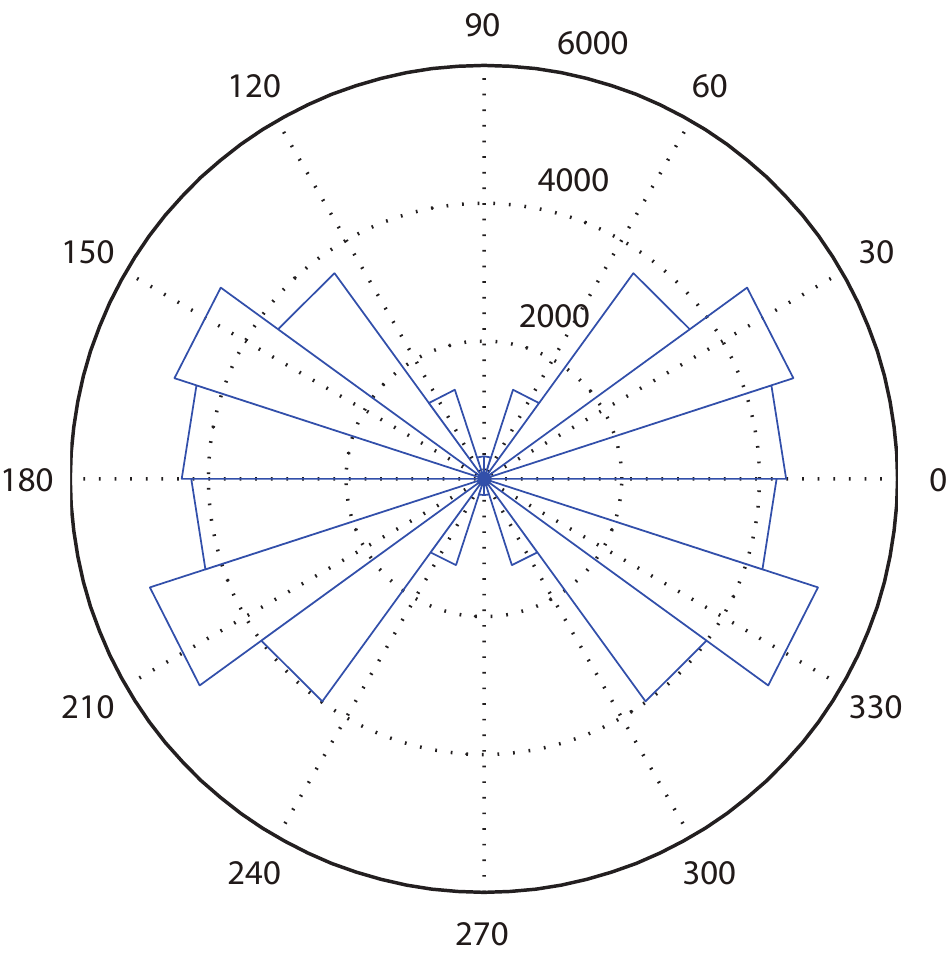}
}
\subfigure[]{
    \includegraphics[width=3.0cm]{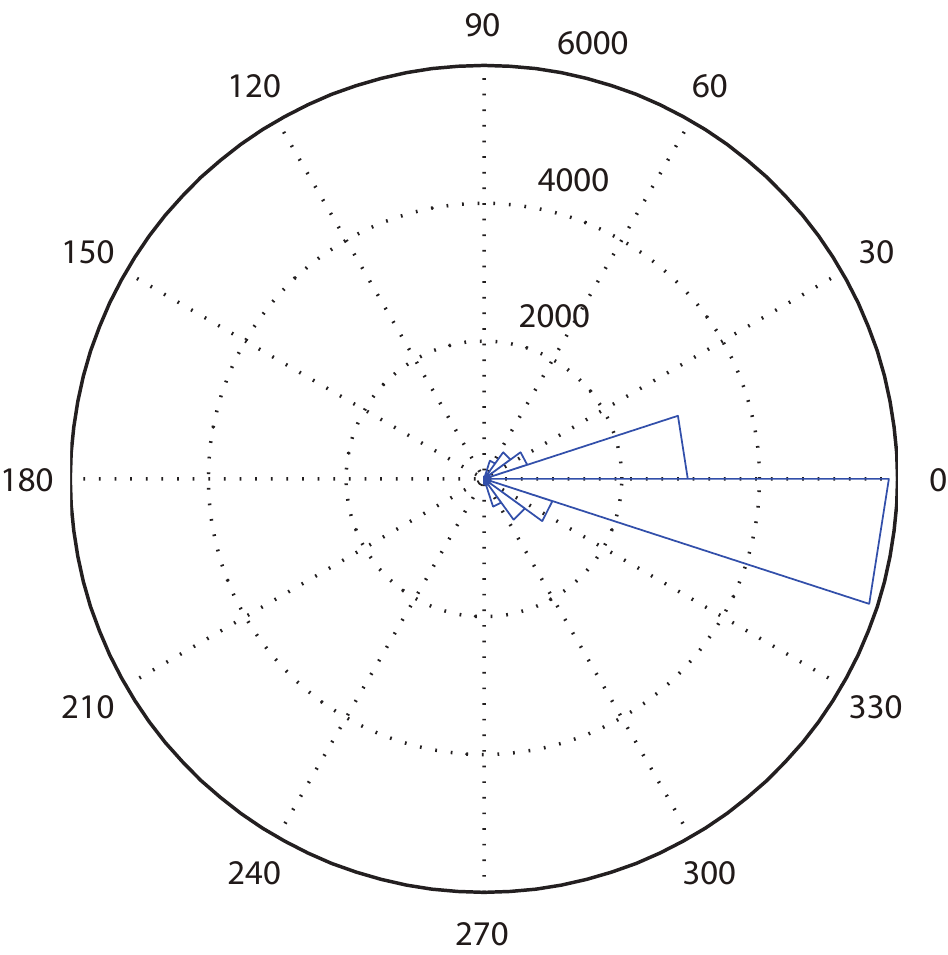}
}
  \end{center}
  \caption{The view distributions of the Multiview-Action3D dataset (left) and MSR-DailyActivity3D dataset (right).}\vspace{-12pt}
  \label{fig:deg_distributions}
\end{figure}

\begin{figure}
  \begin{center}
    \includegraphics[width=8cm]{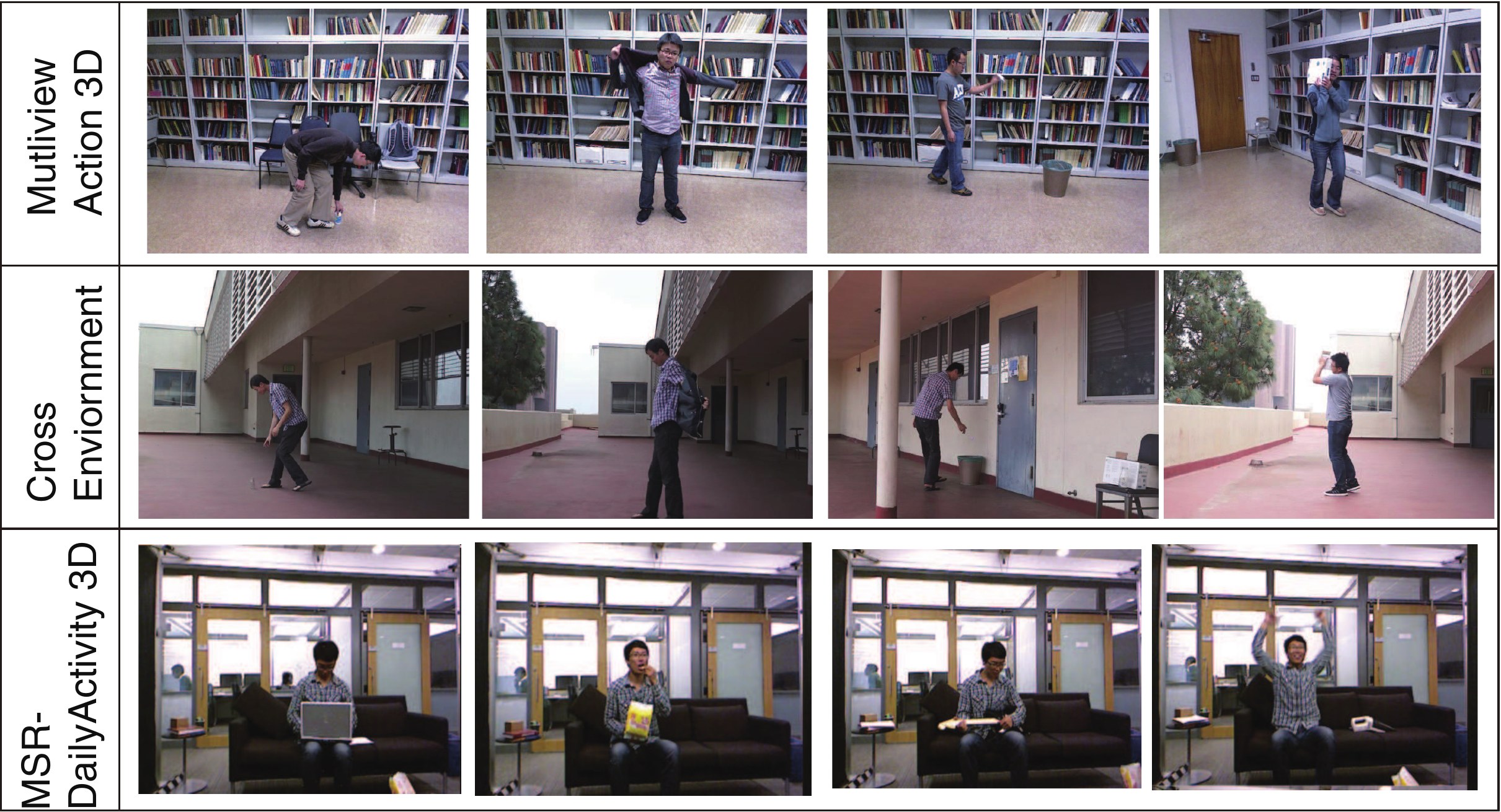}
  \end{center}
  \caption{Sample frames of  Multiview Action3D dataset, cross environment
test data, and MSR-DailyActivity3D dataset~\cite{Wang2012}.}
  \label{fig:examples}
\end{figure}

Northwestern-UCLA Multiview 3D event dataset \footnote{\texttt{http://users.eecs.northwestern.edu/~jwa368/my\_data.html}} contains RGB, depth and human skeleton data captured
simultaneously by three Kinect cameras. This dataset include 10 action
categories: \emph{pick up with one hand}, \emph{pick up with two
  hands}, \emph{drop trash}, \emph{walk around}, \emph{sit down},
\emph{stand up}, \emph{donning}, \emph{doffing}, \emph{throw}, \emph{carry}. Each action is
performed by 10 actors. Fig. \ref{fig:examples} shows some example frames
of this dataset. The view distribution is shown
in Fig. \ref{fig:deg_distributions}. This dataset contains data taken from a variety of viewpoints.

The comparison of the recognition accuracy of the proposed algorithm with the baseline algorithms is shown in Table \ref{tab:multiview}. We
compare with virtual views~\cite{Li2012}, Hankelet~\cite{li2012cross},
Action Bank \cite{Sadanand2012} and Poselet~\cite{Maji2011a}. For
Action Bank, we use the actions provided by \cite{Sadanand2012} as well as a portion of the videos in our dataset as action banks. For Poselet,
we use the Poselets provided by \cite{Maji2011a}. We also compare our model with training one dedicated model for each view,
which is essentially a mixture of deformable part models (DPM),  to compare the robustness of the proposed method under different
viewpoints with DPM model. We have $50$ {\em pose} nodes for all the actions and $10$ child {\em view} nodes for one {\em pose} node for both mixture of DPM and
MST-AOG. The number of the {\em part} nodes in DPM and MST-AOG is both $1320$ (different poses can have different number of parts). MST-AOG also
has 2 child {\em low-resolution feature} nodes for each {\em action} node. These parameters are chosen via cross-validation. In MST-AOG, the appearance/motion
and geometrical relationship of the {\em part} nodes are shared and learned across different {\em view} nodes, but the mixture of DPM treats them
independently.

We perform recognition experiments under three settings.\vspace{-6pt}
\begin{itemize}\addtolength{\itemsep}{-0.5\baselineskip}
\item {\em cross-subject} setting: We use the samples from 9 subjects
as training data, and leave out the samples from 1 subject as testing data.
\item  {\em cross-view} setting: We use the samples
from 2 cameras as training data, and use the samples from 1 camera as testing data.
\item {\em cross-environment}  setting: We apply
the learned model to the same action but captured in a different environment.
Some of the examples of the cross environment testing data are
shown in Fig. \ref{fig:examples}.
\end{itemize}\vspace{-6pt}
These settings can evaluate the robustness to the variations in different subjects, from different views, and
in different environments.

\begin{figure}
  \begin{center}
    \includegraphics[width=8cm]{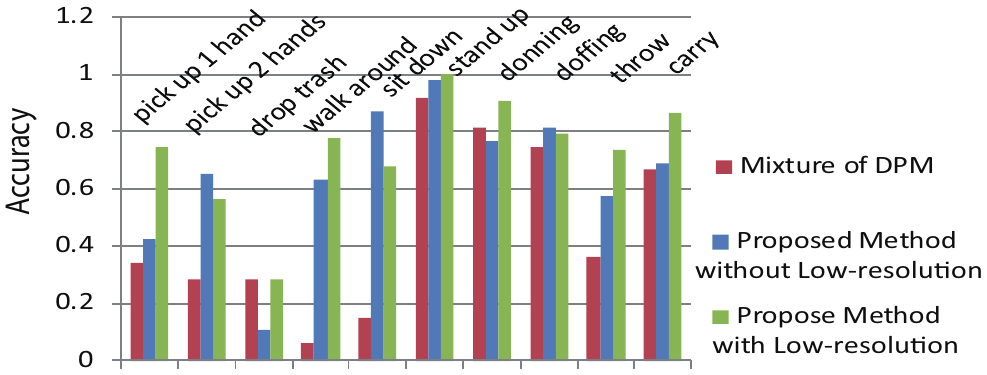}
  \end{center}
  \caption{The recognition accuracy under cross-view setting.}\vspace{-18pt}
  \label{fig:accuracy_comparison}
\end{figure}

The proposed algorithm achieves the best performance under all three settings. Moreover, the proposed method is rather robust under the
cross-view setting.
In contrast, although the state-of-the-art
local-feature-based cross-view action recognition methods~\cite{li2012cross,Li2012} are relatively robust to viewpoint changes, the overall
accuracy of these methods is not very high, because the  local features are not enough to discriminate the subtle differences of the actions in this dataset. Moreover, these methods are sensitive to the changes of the environment.
The Poselet method is robust to environment changes, but it is sensitive to viewpoint changes.
 Since the mixture of DPM does not model the relations across different view, its performance degrades
significantly under cross-view setting. The comparison of the recognition accuracy of the different methods under cross-view setting is shown in
Fig. \ref{fig:accuracy_comparison}.  We also observe that
utilizing low-resolution  features can increase the recognition accuracy, and the proposed method is also robust
under cross environment setting.

\begin{table}
\begin{small}
  \begin{center}
    \begin{tabular}{ | l | c| c|r | }
      \hline
      Method & C-Subject & C-View & C-Env\\
      \hline
      Virtual View  \cite{Li2012} & 0.507& 0.478 & 0.274\\
      Hankelet \cite{li2012cross} & 0.542 & 0.452 & 0.286 \\
      Action Bank \cite{Sadanand2012} &  0.246 & 0.176 & N/A\\
      Poselet \cite{Maji2011a} & 0.549& 0.245& 0.485\\
      \hline
      Mixture of DPM  & 0.748& 0.461 & 0.688
      \\
      MST-AOG w/o Low-S& 0.789 & 0.653 & 0.719\\
      MST-AOG w Low-S& {\bf 0.816} & {\bf 0.733} &
      {\bf 0.793}\\
      \hline
    \end{tabular}
  \end{center}
\end{small}
  \caption{Recognition accuracy on Multiview-3D dataset.}
  \label{tab:multiview}
\end{table}

\begin{figure}
  \begin{center}
    \includegraphics[width=7.1cm]{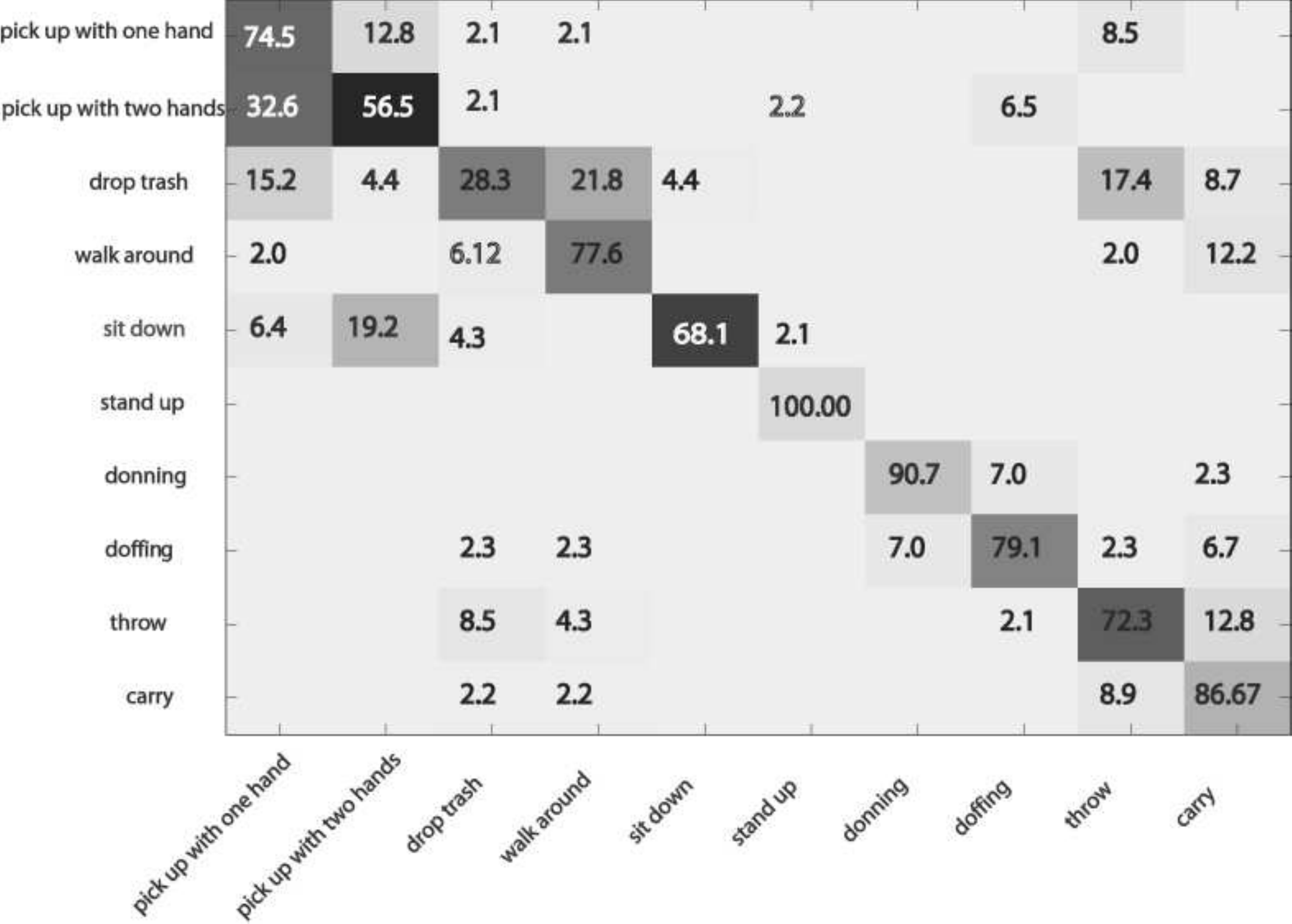}
  \end{center}
  \caption{The confusion matrix of MST-AOG on multiview
    data under cross-view setting (with low-resolution features).}\vspace{-12pt}
  \label{fig:confusion_multivew_crossview}
\end{figure}

The confusion matrix of the proposed methods with low-resolution features under cross-view setting is shown in Fig.~\ref{fig:confusion_multivew_crossview}. The
actions that cause most confusion are ``pick up with one hand'' versus ``pick up with two hands'', because the motion and appearance of these two
actions are very similar. Another action that causes a lot of confusion is ``drop trash'', because the movement of dropping trash can be
extremely subtle for some subjects.

\subsection{MSR-DailyActivity3D Dataset}
The MSR-DailyActivity3D dataset  is a daily activity dataset captured by a Kinect device.
It is a widely used as a Kinect action recognition benchmark.
There are 16 activity types:
\emph{drink}, \emph{eat}, \emph{read book}, \emph{call
  cellphone}, \emph{write on a paper}, \emph{use laptop}, \emph{use
  vacuum cleaner}, \emph{cheer up}, \emph{sit still}, \emph{toss
  paper}, \emph{play game}, \emph{lay down on sofa},
\emph{walk}, \emph{play guitar}, \emph{stand up}, \emph{sit
  down}. If possible, each subject performs an activity in two
different poses: ``sitting on sofa'' and ``standing''. Some example
frames are shown in
Fig.~\ref{fig:examples}. The view distribution of this dataset can be
found in Fig.~\ref{fig:deg_distributions}. Although this dataset is
not a multiview dataset, we compare the performance of the proposed method with the baseline methods
to validate its performance on single view action recognition.

We use the same experimental setting as \cite{Wang2012}, using the samples of half of the subjects as training data, and the samples
of the rest half as testing data. This dataset is very challenging if the 3D skeleton is not used.
The Poselet method~\cite{Maji2011a} achieves 23.75\% accuracy, because many of
the actions in this dataset should be distinguished with motion
information, which is ignored in the Poselet method.  STIP~\cite{Laptev2005} and
Action Bank~\cite{Sadanand2012} do not perform well on this dataset, either.
The proposed MST-AOG method achieves
a recognition accuracy of 73.5\%, which is much better than the baseline methods.

Notice that the accuracy of Actionlet Ensemble method in \cite{Wang2012} achieves 85.5\% accuracy. However, the proposed method only needs one
RGB video as input during testing, while Actionlet Ensemble method requires depth sequences and Kinect skeleton tracking during testing.

\begin{table}
\begin{minipage}{8cm}
  \begin{small}
    \begin{center}
      \begin{tabular}{ | l | r | }
        \hline
        Method & Accuracy \\
        \hline
        STIP \cite{Laptev2005} & 0.545\\
        Action Bank \cite{Sadanand2012} & 0.23\\
        Poselet \cite{Maji2011a}& 0.2375\\
        Actionlet Ensemble \cite{Wang2012} & 0.835\footnote{This result is not directly comparable with MST-AOG, because it uses 3D skeleton.}  \\
        \hline
        MST-AOG & {\bf  0.731}\\
        \hline
      \end{tabular}
    \end{center}
  \end{small}
  \caption{Recognition accuracy for DailyActivity3D dataset.}
\end{minipage}
  \label{tab:DailyActivity}
\end{table}

\begin{figure}
  \begin{center}
    \includegraphics[width=8cm]{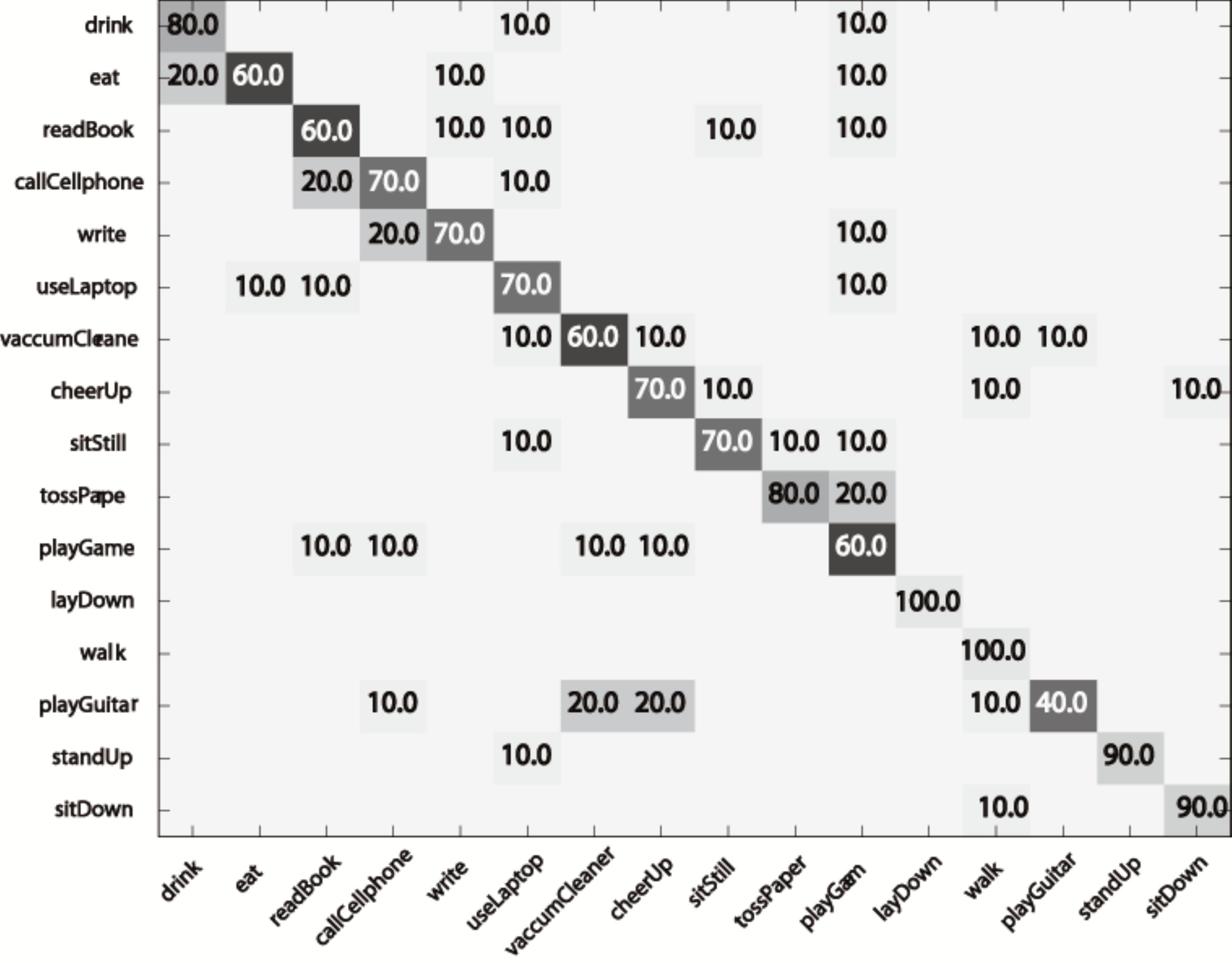}
  \end{center}
  \caption{The confusion matrix of MST-AOG on
    MSR-DailyActivity3D dataset.}
  \label{fig:confusion_daily}
\end{figure}

The confusion matrix of the proposed method on MSR-DailyActivity3D dataset is shown in Fig. \ref{fig:confusion_daily}. We can see that the
proposed algorithm performs well on the actions that are mainly determined by poses or motion, such as ``stand up'', ``sit down'', ``toss
paper'', ``cheer up'', ``call cellphone''. However, recognizing some actions requires us to recognize objects, such as ``playing guitar'' and
``play games''. Modeling the human-object interaction will improve the recognition accuracy for these actions.

\section{Conclusion}
We propose a new cross-view action representation, the MST-AOG model, that can effectively express the geometry, appearance and motion variations
across multiple view points with a hierarchical compositional model. It takes advantage of 3D skeleton data to train, and achieves 2D video
action recognition from unknown views. Our extensive experiments have demonstrated that MST-AOG significantly improves the accuracy and
robustness for cross-view, cross-subject and cross-environment action recognition. The proposed MST-AOG can also be employed to detect the view
and locations of the actions and poses. This will be our future work.

\subsubsection{Acknowledgement}
{\footnotesize
This work was supported in part by DARPA Award FA 8650-11-1-7149, National Science Foundation grant IIS-0916607, IIS-1217302,
and MURI grant ONR N00014-10-1-0933.
Part of the work was done when the first author was visiting VCLA lab in UCLA.}
{\scriptsize
\bibliographystyle{ieee}
\bibliography{../library_handedited}
}

\end{document}